\documentclass[nonacm,sigplan,authorformat=flat]{acmart}

\newcommand{\loopconversion}[0]{Kernel looping}
\newcommand{\ts}[0]{SN40L}

\usepackage{tabularray}
\usepackage{subcaption}
\usepackage[]{hyperref}
\usepackage{tikz}
\usepackage{breqn}
\usepackage{algorithm2e}
\usepackage{listings}
\usepackage{multirow}
\usepackage{booktabs}

\tikzset{white border/.style={preaction={draw,white,line width=4pt}}}

\author{David Koeplinger}
\affiliation{%
  \institution{\textbf{SambaNova Systems, Inc}}
   \country{}
}
\email{david.koeplinger@sambanova.ai}

\author{Darshan Gandhi}
\affiliation{%
  \institution{\textbf{SambaNova Systems, Inc}}
   \country{}  
}
\email{darshan.gandhi@sambanova.ai}

\author{Pushkar Nandkar}
\affiliation{%
  \institution{\textbf{SambaNova Systems, Inc}}
   \country{}  
}
\email{pushkar.nandkar@sambanova.ai}

\author{Nathan Sheeley}
\affiliation{%
  \institution{\textbf{SambaNova Systems, Inc}}
   \country{}  
}
\email{nathan.sheeley@sambanova.ai}

\author{Matheen Musaddiq}
\affiliation{%
  \institution{\textbf{SambaNova Systems, Inc}}
   \country{}  
}
\email{matheen.musaddiq@sambanova.ai}

\author{Leon Zhang}
\affiliation{%
  \institution{\textbf{SambaNova Systems, Inc}}
   \country{}  
}
\email{leon.zhang@sambanova.ai}

\author{Reid Goodbar}
\affiliation{%
  \institution{\textbf{SambaNova Systems, Inc}}
   \country{}  
}
\email{reid.goodbar@sambanova.ai}

\author{Matthew Shaffer}
\affiliation{%
  \institution{\textbf{SambaNova Systems, Inc}}
   \country{}  
}
\email{matthew.shaffer@sambanova.ai}

\author{Han Wang}
\affiliation{%
  \institution{\textbf{SambaNova Systems, Inc}}
   \country{}  
}
\email{han.wang@sambanova.ai}

\author{Angela Wang}
\affiliation{%
  \institution{\textbf{SambaNova Systems, Inc}}
   \country{}  
}
\email{angela.wang@sambanova.ai}

\author{Mingran Wang}
\affiliation{%
  \institution{\textbf{SambaNova Systems, Inc}}
   \country{}  
}
\email{mingran.wang@sambanova.ai}

\author{Raghu Prabhakar}
\affiliation{%
  \institution{\textbf{SambaNova Systems, Inc}}
  \country{}
}
\email{raghu.prabhakar@sambanova.ai}

\begin{document}

\title{Kernel Looping: Eliminating Synchronization Boundaries for Peak Inference Performance}


\begin{abstract}
Token generation speed is critical to power the next wave of AI inference applications. GPUs significantly underperform during token generation due to synchronization overheads at kernel  boundaries, utilizing only 21\% of their peak memory bandwidth. While recent dataflow architectures mitigate these overheads by enabling aggressive fusion of decoder layers into a single kernel, they too leave performance on the table due to synchronization penalties at layer boundaries.

This paper presents \emph{kernel looping}, a specialized global optimization technique which exploits an optimization opportunity brought by combining the unique layer-level fusion possible in modern dataflow architectures with the repeated layer structure found in language models. Kernel looping eliminates synchronization costs between consecutive calls to the same kernel by transforming these calls into a single call to a modified kernel containing a pipelined outer loop. We evaluate kernel looping on the SambaNova \ts \space Reconfigurable Dataflow Unit (RDU), a commercial dataflow accelerator for AI. Experiments demonstrate that kernel looping speeds up the decode phase of a wide array of powerful open-source models by up to 2.2$\times$ on \ts. Kernel looping allows scaling of decode performance over multiple \ts \space sockets, achieving speedups of up to 2.5$\times$. Finally, kernel looping enables \ts \space to achieve over 90\% of peak performance on 8 and 16 sockets and achieve a speedup of up to 3.7$\times$ over DGX H100. Kernel looping, as well as the models evaluated in this paper, are deployed in production in a commercial AI inference cloud.



\end{abstract}

\maketitle 
\pagestyle{plain} 

\section{Introduction}

\begin{table*}
\centering\small
\begin{tabular}{|p{1.2in} |p{0.3in}|p{0.5in}|p{0.7in}|p{0.65in}|p{0.54in}|p{0.45in}|p{0.35in}|p{1in}|} \hline
\textbf{Model, Precision} & \textbf{Batch Size} & \textbf{Context Length} & \textbf{Parameters (GB)} & \textbf{KV Cache (GB)} & \textbf{GFLOPs / token} & \textbf{GBytes / token} & \textbf{FLOPs / byte} & \textbf{Memory-bound on DGX H100?} \\ \hline
Llama 3.1 8B, BF16   & 1     & 4096    & 14.9  & 0.5   & 20     & 15.4    & 1.3  & \textbf{YES} \\ 
                     & 1     & 65536   & 14.9  & 8     & 50     & 22.9      & 2.2  & \textbf{YES} \\ 
                     & 32    & 4096    & 14.9  & 16    & 640    & 30.9      & 20.7   & \textbf{YES} \\                    
                     & 32    & 65536   & 14.9  & 256   & 1600   & 270.9     & 5.9  & \textbf{YES} \\ \hline 
Llama 3.1 70B, BF16  & 1     & 4096    & 130.4 & 1.25  & 150    & 131.6  & 1.1  & \textbf{YES} \\ 
                     & 1     & 65536   & 130.4 & 20    & 310    & 150.4     & 2.1  & \textbf{YES} \\ 
                     & 32    & 4096    & 130.4 & 40    & 4800   & 170.4     & 28.2 & \textbf{YES} \\                    
                     & 32    & 65536   & 130.4 & 640   & 9920   & 770.4     & 12.9 & \textbf{YES} \\ \hline   
Llama 3.1 405B, BF16 & 1     & 4096    & 754.4 & 1.96  & 840    & 756.3  & 1.1  & \textbf{YES} \\ 
                     & 1     & 65536   & 754.4 & 31.5  & 1340   & 785.9   & 1.7  & \textbf{YES} \\ 
                     & 32    & 4096    & 754.4 & 62.7  & 26880  & 817.1  & 32.9 & \textbf{YES} \\                    
                     & 32    & 65536   & 754.4 & 1008  & 42880  & 1762.4    & 24.3 & \textbf{YES} \\ \hline                         
\end{tabular}
\caption{Total operations, bytes, and operation intensities for the decode phase of Llama3.1 models at various batch sizes and context lengths. Any operation intensity less than 333 is memory bandwidth-bound on DGX H100.}
\label{tab:op-intensity}
\end{table*}

\begin{figure}
     \begin{subfigure}[b]{0.8\linewidth}
        \centering
         \includegraphics[width=\textwidth]{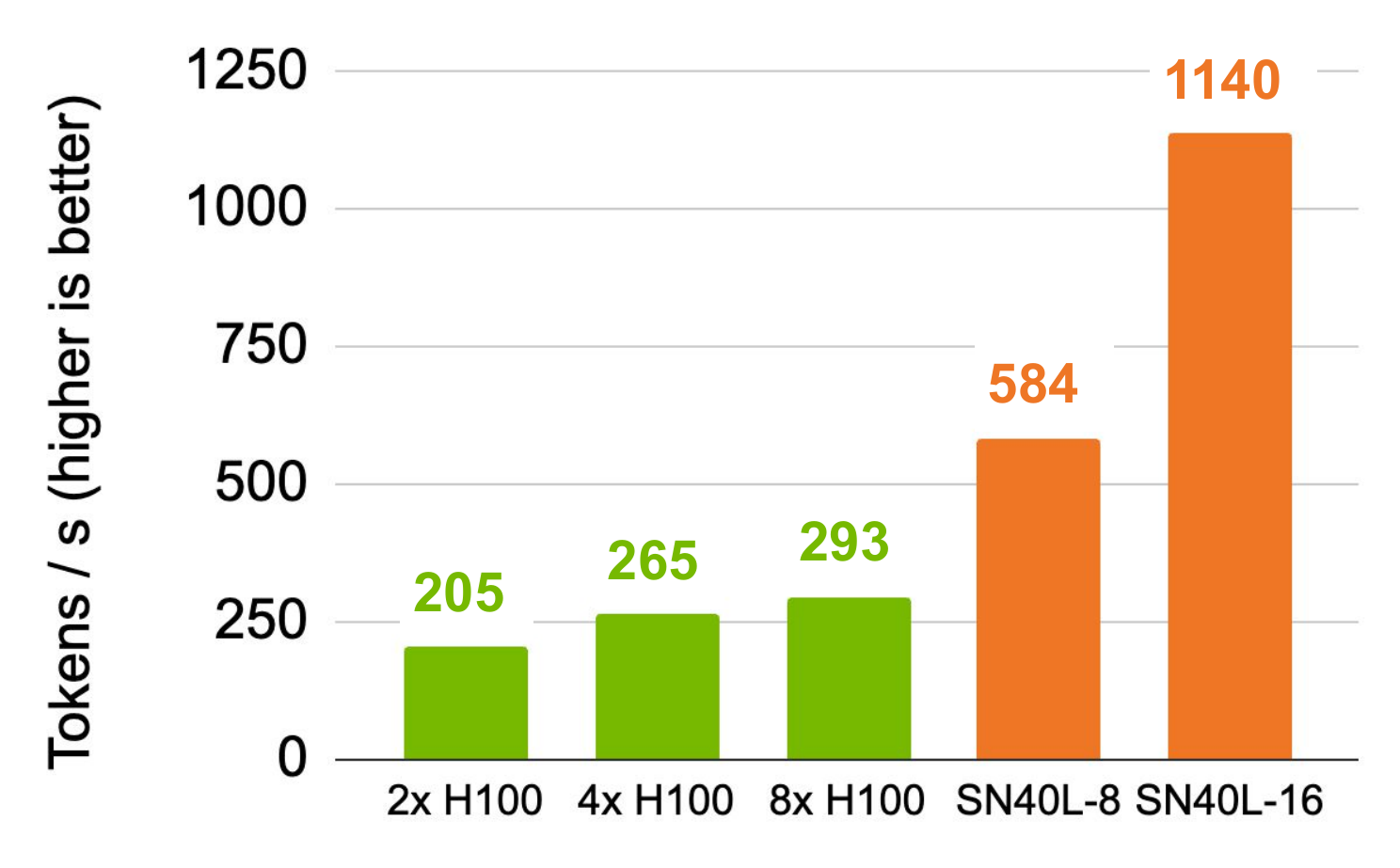}
         \caption{8B, BS=1, Decoding speed in tokens/second.}
         \label{fig:rdu-vs-h100-8b-bs1-tps}
     \end{subfigure}
     \begin{subfigure}[b]{0.8\linewidth}
        \centering
         \includegraphics[width=\textwidth]{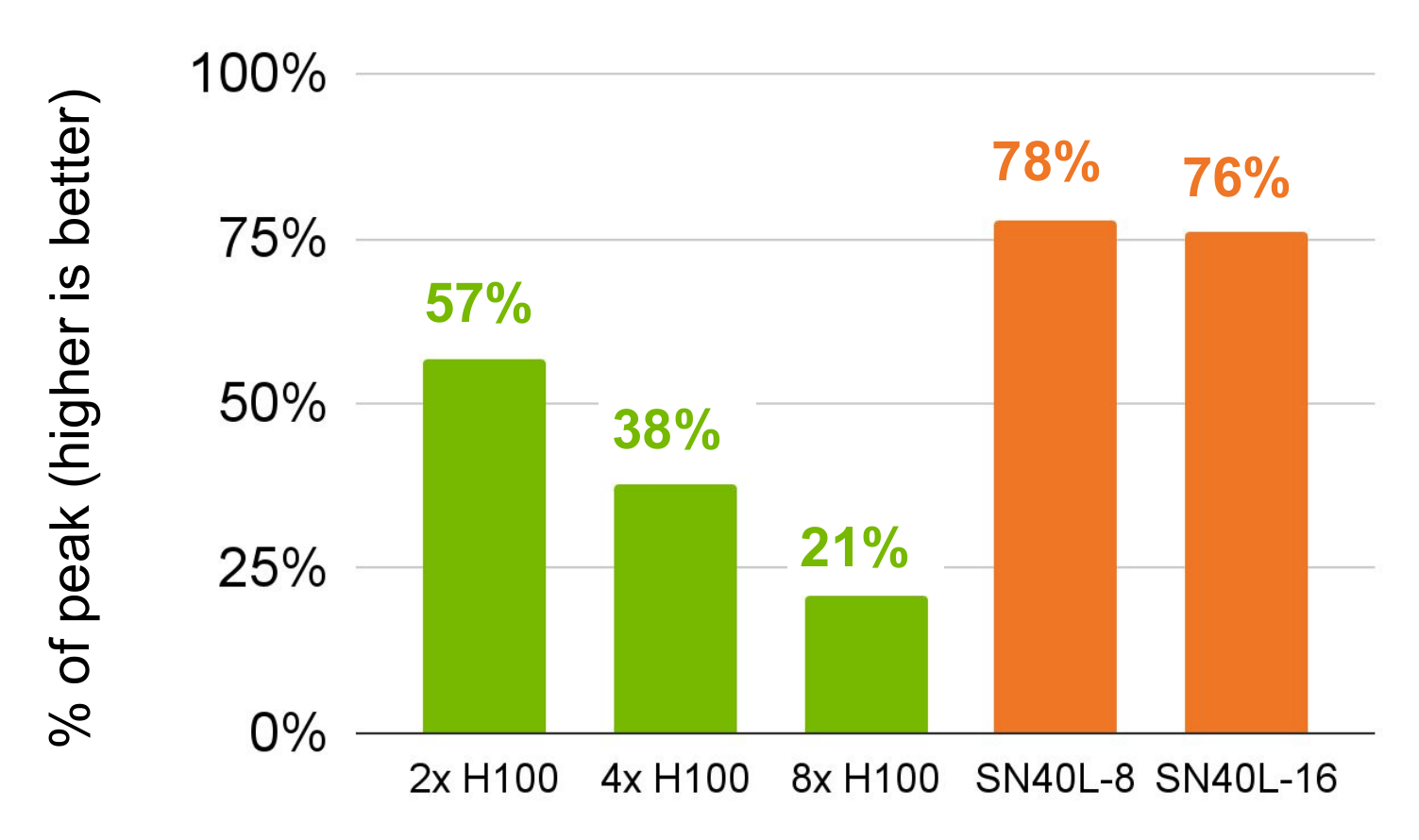}
         \caption{8B, BS=1, Percentage of peak performance.}
         \label{fig:rdu-vs-h100-8b-bs1-util}
     \end{subfigure}
        \caption{Token generation performance of Llama3.1-8B on DGX H100 and \ts. Synchronization overheads at kernel call boundaries in DGX H100 inhibit utilizing HBM bandwidth efficiently. Kernel looping (two orange bars to the right) mitigates these overheads on \ts, and have been deployed in production.
        }
        \label{fig:gpu-sucks-intro}
\end{figure}

In recent years, the landscape of open-source language models has evolved significantly, with state-of-the-art (SOTA) models becoming smaller while achieving very high accuracy~\cite{llama-herd, qwen2, qwen2.5, mixtral, deepseekv2}. These compact models, often pre-trained or distilled from larger ones over vast datasets~\cite{llama-herd}, have achieved remarkable accuracy through advancements in model architecture and training techniques. As model context length continues to grow (sometimes exceeding 128,000 tokens~\cite{llama-herd}), more sophisticated prompt engineering and in-context learning techniques~\cite{gpt-icl} have emerged, shifting focus from training to inference. 

Inference involves generating a sequence of \emph{output} tokens, given a sequence of \emph{input} tokens. Output tokens are generated in two distinct phases: ``prefill'' and ``decode''. Prefill is the process of producing the first output token, while decode is the process of producing all subsequent output tokens in an autoregressive manner. The time spent in the prefill phase is called Time to First Token (TTFT). The time taken to produce one token in the decode phase is called Time per Output Token (TPOT). Decoding performance is often reported as a throughput: tokens per second~\cite{aa}. The prefill phase is often compute-bound on modern AI accelerators~\cite{agrawal2024taming} due to the high operational intensity in its compute graph. Conversely, the decode phase is memory-bound due to low operational intensity as shown in Table~\ref{tab:op-intensity}. The ratio of input:output tokens also varies widely based on the task. For instance, summarization tasks have a high input:output token ratio, while math and coding questions can have a low input:output token ratio.  

With access to powerful open-source models, the machine learning community has produced advanced inference techniques that drastically increase the number of output tokens for an input prompt. Chain-of-Thought (CoT) prompting~\cite{cot, reflexion, cove} performs complex reasoning tasks by breaking down problems into a series of logical steps. However, this method generates multiple \emph{intermediate tokens} during inference, significantly increasing the overall time spent in the decode phase. Similarly, test-time scaling techniques~\cite{testtimescaling1, testtimescaling2, testtimescaling3} have demonstrated that inference efficiency can be improved by increasing the output tokens generated, leveraging the model's ability to iterate and refine its outputs. Such iterative methods rely on a large context window that holds all tokens generated at each step. While a naive deployment of such solutions can negatively impact TTFT, innovations like context caching~\cite{context-caching-1, context-caching-2, context-caching-3} mitigate this impact by preserving KV cache values across multiple turns. Context caching accelerates prefill by reducing the  compute required during this phase in multi-turn conversations. With context caching, optimization needs shift back to the decode phase. This brings optimizations like speculative decoding~\cite{spec-decoding, specinfer} to the forefront. Speculative decoding relies on smaller draft models to have high decode throughput to be effective. This way, the smaller model generates tokens rapidly which are then checked and corrected by a larger model at a chosen cadence~\cite{spec-decoding}. Consequently, fast token generation on both large and small models is the most important enabler for modern inference.

The decode phase of inference is memory-bound on modern AI accelerators like the DGX H100~\cite{h100-gpu} due to its low operational intensity. Table~\ref{tab:op-intensity} shows the operations, bytes, and the operation intensities to generate one token from various Llama3.1 models under different batch sizes and sequence lengths. In this table, we estimate the FLOPs/byte by assuming that generating a single token requires reading all parameters and KV cache values once. The reported compute FLOPs are obtained by adding all matrix multiply operations in all layers. Streaming operations like softmax and RMSNorm account for a much smaller fraction of the compute and do not change the overall insight below. DGX H100 has a peak BF16/FP16 compute capability of 1000 TFLOPs (excluding structured sparsity), and a peak HBM bandwidth of 3 TB/s~\cite{dgx-h100}. Any configuration with an operation intensity less than $1000 / 3 = 333$ will therefore be memory bandwidth-bound. As shown in Table~\ref{tab:op-intensity}, all Llama 3.1 model configurations are memory bandwidth-bound. Note that for the combination of large sequence length and large batch size, the FLOPs/byte goes \emph{down} as sequence length increases. Unlike batch size, which linearly increases both compute FLOPs and KV cache size, sequence length only impacts a subset of the operators during decode. Consequently, the KV cache size grows faster than the compute FLOPS, leading to lower operational intensity.

Batching is a common technique to increase operation intensity and improve the overall Performance/TCO. However, batch sizes cannot be made arbitrarily large. Both prefill TFLOPs (which is already compute-bound) and KV cache size increase linearly with batch size. Consequently, higher batch sizes increase TTFT which impacts the end-user SLA and interactivity. Larger KV cache sizes can exceed the limited HBM capacity available on accelerators. Finally, the number of output tokens produced can widely vary for each query in the batch. This variability in output tokens can utilize hardware inefficiently and increase overall queuing delays. As a result, many inference applications, such as chatbots or LLM agents, often limit batch size to ensure a reasonable turnaround time for user queries. ~\cite{liu2024andes}~\cite{zhong2024distserve}.

Figure~\ref{fig:gpu-sucks-intro} plots the decode speed for Llama 3.1 8B with TensorRT-LLM~\cite{tensorrt-llm} on 2, 4, and 8 H100 GPUs (green bars). The last two orange bars, in comparison, are the contribution of this paper. In Figure~\ref{fig:rdu-vs-h100-8b-bs1-util}, a value of 100\% corresponds to the performance obtained if the available HBM bandwidth were fully utilized to transfer weights and the KV cache from memory, with everything else happening in its shadow. We can observe that GPUs operate well below their roofline performance for decode. Furthermore, decode performance scales poorly with increasing numbers of GPUs. Much of this inefficiency stems from \textbf{\emph{forced synchronization}} overheads at kernel call boundaries. Rigidity in the GPU's on-chip memory and interconnect, limited on-chip SRAM, and programming model constraints limit the number of operators that can be fused and executed asynchronously in a kernel~\cite{sn40l}.

Reconfigurable Dataflow Architectures (RDAs) have recently emerged as promising commercial alternatives to GPUs~\cite{sn40l}. Dataflow architectures provide a much more flexible hardware substrate that enables ``layer-level fusion'', or fusing entire layers into a single kernel call~\cite{sn40l, sn40l-hotchips}, significantly reducing the number of kernel calls compared to GPUs. However, we quantify and argue that \textbf{\emph{dataflow architectures incur synchronization overheads at kernel call boundaries too}}. Furthermore, synchronization costs can be nontrivial: reprogramming a dataflow architecture at kernel boundaries involves loading a new configuration bitstream onto all on-chip units, which is a heavier operation than a typical instruction fetch. These synchronization overheads can quickly wipe out the performance upside from higher operator fusion. Note that synchronization points at kernel call boundaries are not fundamental to the algorithm itself. Rather, it is imposed by the limitations of the underlying hardware platform and/or programming model.

This paper proposes a compiler optimization to eliminate unnecessary synchronization boundaries. We build on the benefits of layer-level fusion in dataflow architectures and exploit repetitive layers found in the most powerful model architectures to date. We propose \emph{kernel looping}, a global optimization technique that rewrites groups of repeated kernel calls into a single kernel call with a pipelined outer loop. 
Kernel looping has two key benefits: (i) eliminating synchronization between calls to the same kernel, and (ii) overlapping the compute and communication across kernels. We evaluate the benefits of kernel looping on the SambaNova \ts \space Reconfigurable Dataflow Unit (RDU)~\cite{sn40l, sn40l-hotchips}, a commercial RDA amenable to aggressive layer-level fusion. Figure~\ref{fig:gpu-sucks-intro} shows that with kernel looping, Llama3.1 8B can run at 78\% of roofline performance on the  \ts, outperforming DGX H100 by 2$\times$. Kernel looping also helps scale performance across more \ts \space sockets, achieving 76\% of roofline performance on 16 sockets. Kernel looping and all models reported in this paper have been deployed in production, and are currently in use by several customers and developers.

This paper makes the following key contributions:
\begin{itemize}
    \item In Section~\ref{sec:motivation}, we provide detailed performance characterization of token generation performance on NVIDIA DGX H100 to show that H100 operates under 25\% of its peak roofline performance, largely due to synchronization overheads at kernel call boundaries.
    \item We quantify that synchronization overheads on dataflow architectures like \ts \space account for 30\% of overall token generation time on Llama3.1-8B, and can wash away the benefits of layer-level fusion. We describe an optimization opportunity exposed by layer-level fusion which can mitigate these overheads.
    \item In Section~\ref{sec:system}, we introduce kernel looping, a compiler optimization that automatically rewrites a chain of calls to the same kernel into a single kernel call.
    \item In Section~\ref{sec:evaluation}, we demonstrate the generality, scalability, and performance improvement of kernel looping over a wide array of powerful open-source language models, and show a 1.6x geometric mean speedup on \ts \space without kernel looping. We show that with kernel looping, \ts \space achieves a speedup of up to 3.7$\times$ over DGX H100, and can generate tokens at over 90\% efficiency.
\end{itemize}


\section{Decoder Performance Characterization}
\label{sec:motivation}
\begin{figure*}
        \centering
         \includegraphics[width=0.9\linewidth]{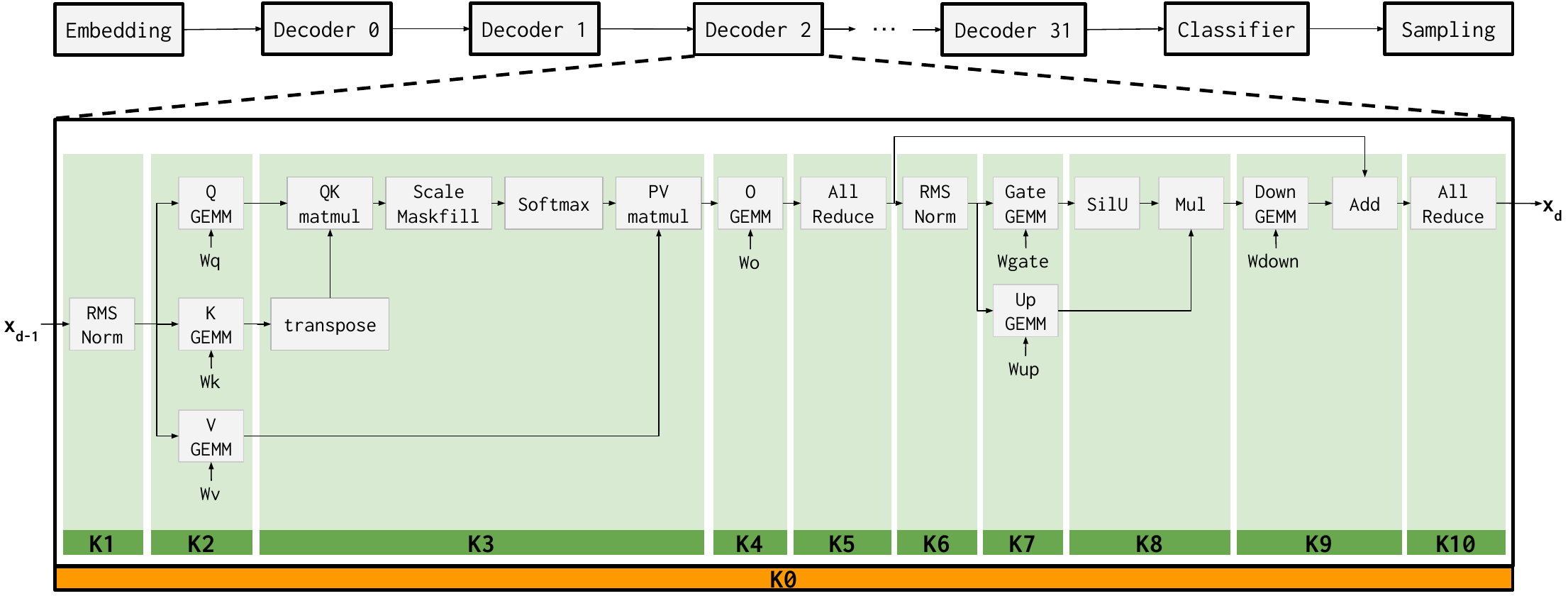}
         \caption{Llama3.1-8B with 32 decoder layers, and the operators within one decoder layer. Labels $K1$ -- $K10$ (green) are kernel calls on the DGX H100, obtained by profiling the execution of Llama3.1-8B using TensorRT-LLM~\cite{tensorrt-llm} with the NVIDIA Nsight profiler~\cite{nsight-systems}. Label $K0$ (orange) corresponds to one kernel call on \ts \space for the entire decoder layer. Collective communication operations like \emph{allreduce} do not execute asynchronously with other operators on the DGX H100. On \ts, \emph{allreduce} is pipelined and overlapped with other operators. }
         \label{fig:decoder-kernels}
\end{figure*}

\begin{figure*}
        \centering
         \includegraphics[width=0.92\linewidth]{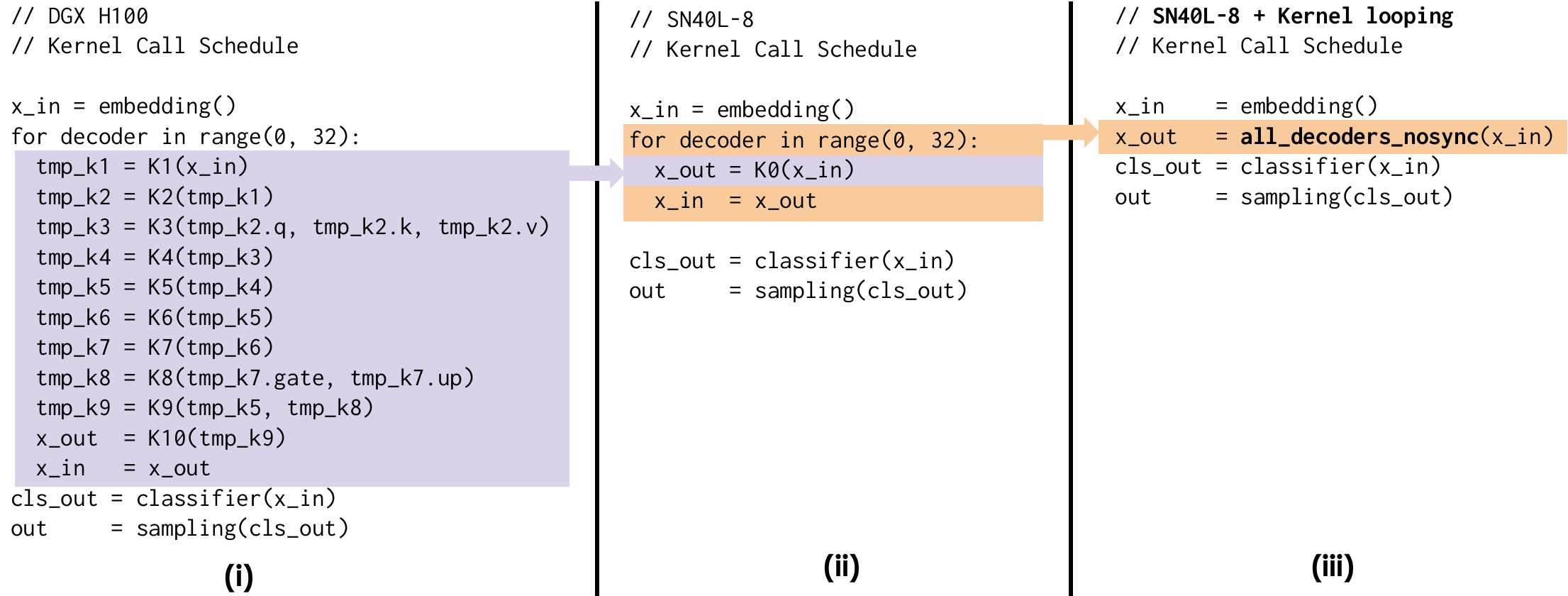}
         \caption{Pseudocode showing kernel call schedule to generate one output token on (i) DGX H100 with TensorRT-LLM~\cite{tensorrt-llm}, (ii) \ts-8 with layer fusion, and (iii) this work. On DGX H100, a single output token requires 323 kernel calls, while \ts-8 needs 35 kernel calls. In this work, we reduce the number of kernel calls per output token on \ts \space to 4.}
         \label{fig:decoder-schedule}
\end{figure*}

In this section, we use Llama3.1-8B as a driving example to characterize token generation performance on two platforms: DGX H100~\cite{dgx-h100} and \ts. Here, \ts-8 corresponds to a system with 8 \ts \space sockets, similar to DGX H100 with 8 GPUs. Specifically, we highlight and quantify the inefficiencies on both platforms due to synchronization at kernel call boundaries, thus motivating the need for the techniques described in Section~\ref{sec:system}. We chose Llama3.1-8B because it is a very popular open-source model. But more importantly, Llama3.1-8B plays an important role as a draft model to accelerate larger models like Llama3.1-70B and Llama3.1-405B using speculative decoding~\cite{spec-decoding}. While we use a specific model to illustrate  concepts, the insights apply broadly to a larger variety of model architectures.

Figure~\ref{fig:decoder-kernels} and Figure~\ref{fig:decoder-schedule} depict the structure of Llama3.1-8B and its execution on both DGX H100 and \ts. Figure~\ref{fig:decoder-kernels} shows that Llama3.1-8B has 4 major types of blocks: one \texttt{Embedding} layer, 32 \texttt{Decoder} layers, one \texttt{Classifier} layer, and \texttt{Sampling}. While \texttt{Sampling} is not part of the model, we include it here as it is executed once for every output token, similar to the other layers. Figure~\ref{fig:decoder-kernels} also shows the key operators comprising a decoder. As decoder layers dominate token generation time, the rest of this section will focus on how a single decoder layer is executed on multiple sockets.

\begin{figure}
     \begin{subfigure}[b]{0.8\linewidth}
        \centering
         \includegraphics[width=\textwidth]{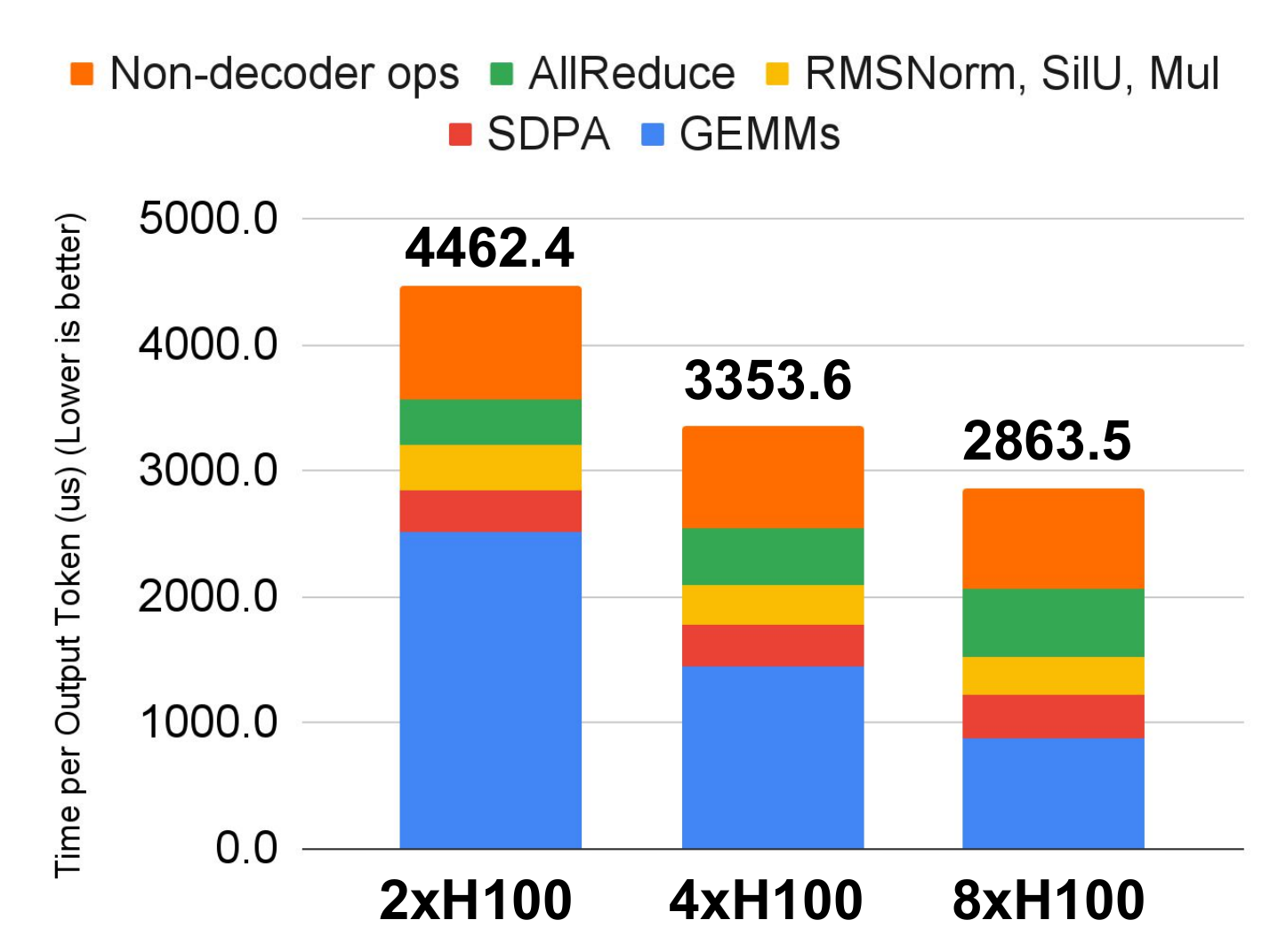}
         \caption{DGX H100 raw TPOT breakdown.}
         \label{fig:h100-breakdown-raw}
     \end{subfigure}
     \begin{subfigure}[b]{0.8\linewidth}
        \centering
         \includegraphics[width=\textwidth]{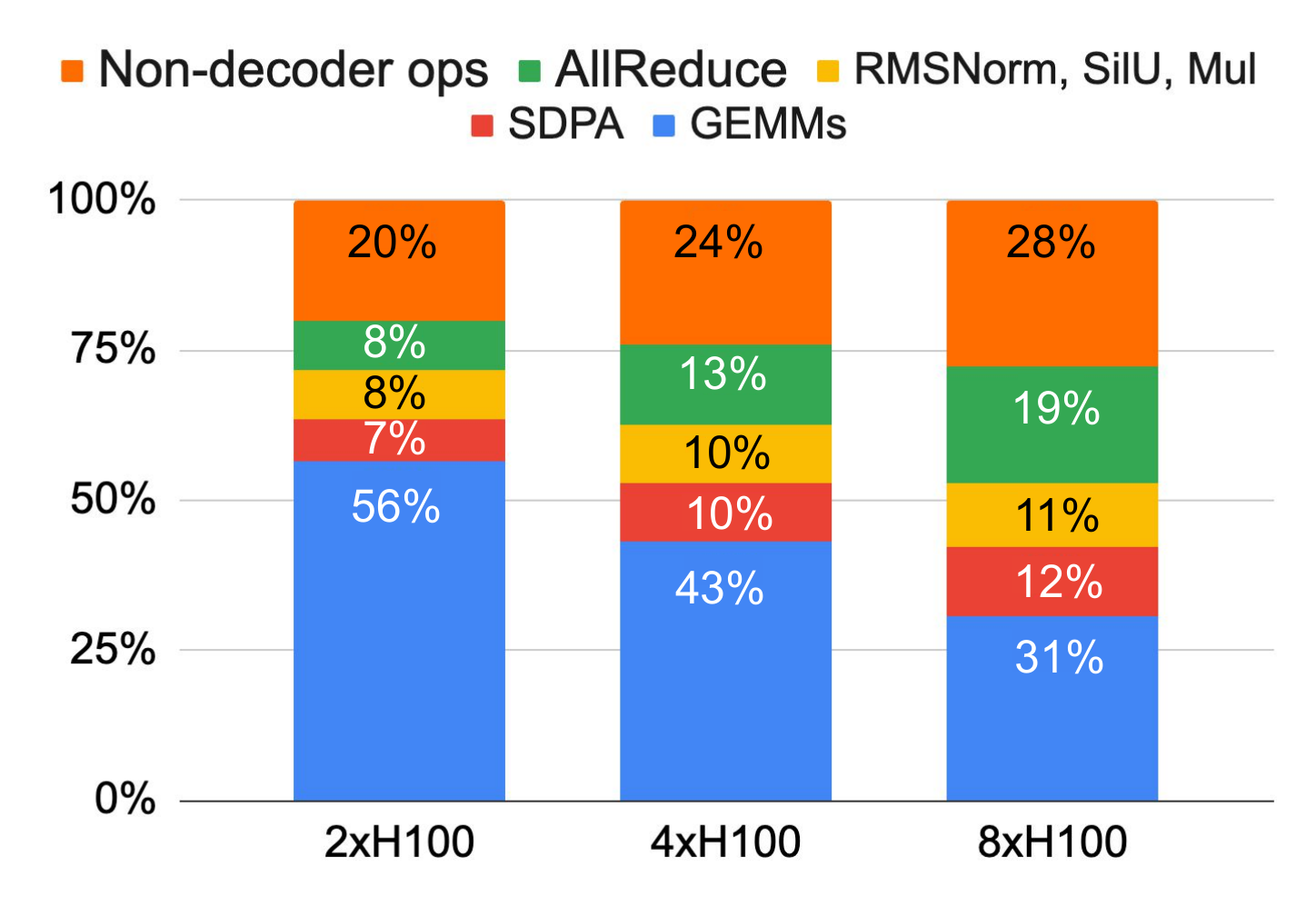}
         \caption{DGX H100 TPOT breakdown by percentage. }
         \label{fig:h100-breakdown-percent}
     \end{subfigure}
         \caption{Time-per-Output-Token (TPOT) breakdown for Llama3.1-8B on 2, 4, and 8 H100s. GEMMs scale better than other operators (although not ideally). \emph{Allreduce} and other operators demonstrate poor scaling, thus accounting for a larger fraction of the overall token generation time.
        }
        \label{fig:h100-breakdown}
\end{figure}

Each decoder layer is parallelized across multiple sockets to extract higher performance. In a DGX H100, a decoder is parallelized across 8 H100 GPUs. While several parallelization strategies exist in literature, we study the Tensor-Parallel (TP) mapping here as it is a commonly deployed mechanism. TP mapping involves \emph{sharding} weights and input tensors along one or more dimensions, and distributing each shard to a different socket. Sharding introduces distributed reduction and broadcast operations across sockets like \emph{allreduce}.

Figure~\ref{fig:decoder-kernels} shows the execution of a decoder on DGX H100 and \ts-8. On both platforms, a ``kernel'' is a unit of work dispatched to the accelerator, which executes one or more operators in a fused fashion. Kernels act as forced synchronization barriers on both platforms. Specifically, operators belonging to different kernels are not executed asynchronously. For a TP-mapped kernel, a kernel call boundary executes an implicit synchronization barrier across all sockets. The key difference between the two platforms is the \emph{number} of operators fused into a single kernel call. On H100, a decoder is broken up and executed as 10 kernel calls, labeled in green in Figure~\ref{fig:decoder-kernels} as $K1$ through $K10$. In contrast, the \ts-8 executes \emph{all} operators in the decoder layer as a single kernel call, labeled in orange in Figure~\ref{fig:decoder-kernels} as $K0$. 

The kernel makeup for H100 is obtained by using TensorRT-LLM~\cite{tensorrt-llm} to run Llama3.1-8B and by extracting kernel call information using the NVIDIA Nsight Systems profiler~\cite{nsight-systems}. The kernel makeup on the \ts-8 is similar to prior literature~\cite{sn40l-hotchips} performing layer-level fusion. Collective communication operators introduced by TP mapping like \emph{allreduce} are shown explicitly as a separate operator in Figure~\ref{fig:decoder-kernels}. On H100, \emph{allreduce} is implemented by calling into the NVIDIA Collective Communications Library~\cite{nccl}. On the \ts, \emph{allreduce} is implemented as a cross-chip dataflow operator using the chip-to-chip peer-to-peer (P2P) protocol.

We first characterize the performance of a decoder layer on DGX H100. Profiling data is obtained using benchmarking scripts included with TensorRT-LLM after a few warm up iterations, as recommended by the documentation~\cite{tensorrt-llm}. Using this methodology, we observe that Llama3.1-8B generates tokens at over 300 tokens/s on 8xH100. This is competitive with the optimized implementations of several commercial API providers~\cite{aa}, suggesting that our methodology is representative of a real world deployment that includes key GPU optimizations. Figure~\ref{fig:decoder-schedule}(i) shows the kernel schedule for a single decoder on DGX H100, comprising $K1$ through $K10$. 

Figure~\ref{fig:h100-breakdown} shows the time breakdown of the Time Per Output Token (TPOT) for Llama3.1-8B on 2, 4, and 8 H100 sockets. All kernels studied in this scenario are memory bandwidth-bound. From Figure~\ref{fig:h100-breakdown}, we see that the performance of all kernels does not scale by the same amount with increasing socket count. GEMM performance scales better than other operators, but not linearly. Operators like RMSNorm, SiLU, and SDPA perform at a significantly lower efficiency from their theoretical peak despite having more aggregate HBM bandwidth, indicating that kernel warm up and synchronization costs inhibit scaling. The cost of \emph{allreduce} and other non-decoder operators account for a larger portion of the overall time per output token for 8xH100. 

\begin{figure}
     \begin{subfigure}[b]{0.8\linewidth}
        \centering
         \includegraphics[width=\textwidth]{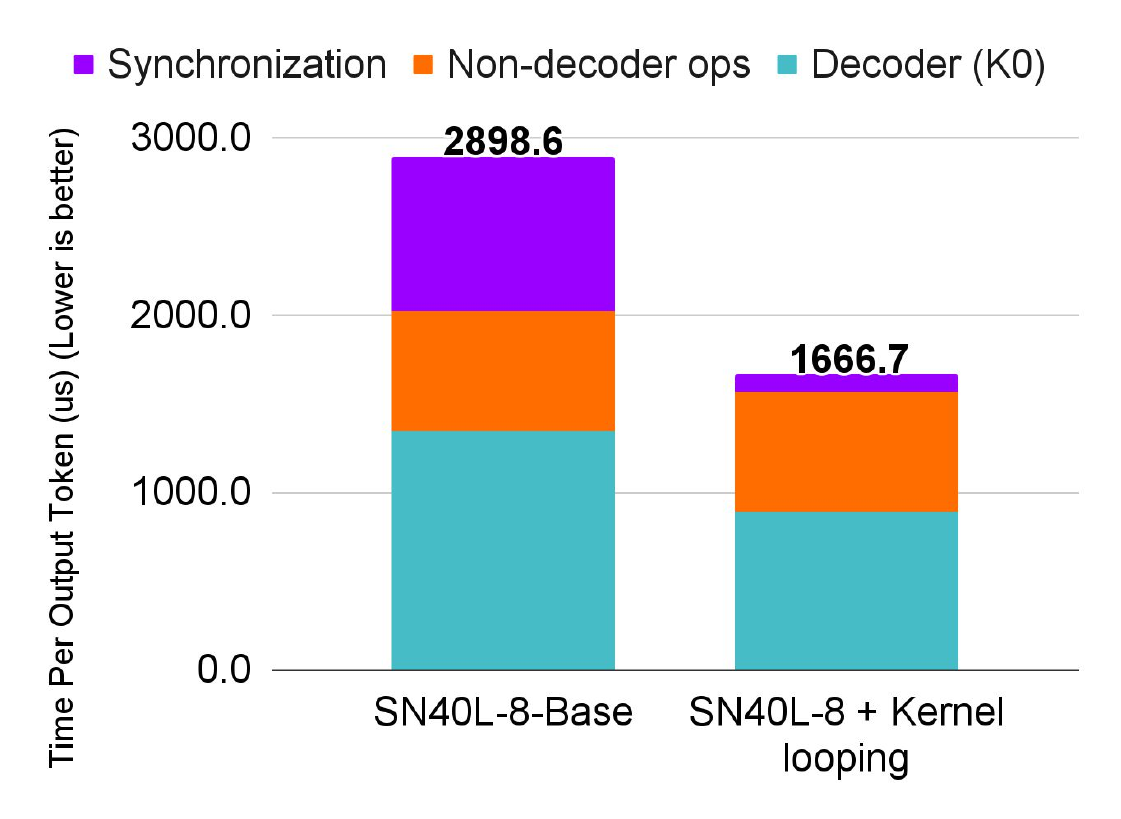}
         \vspace{-15pt}
         \caption{\ts-8 raw TPOT breakdown.}
         \label{fig:sn40l-breakdown-raw}
     \end{subfigure}
     \begin{subfigure}[b]{0.8\linewidth}
        \centering
         \includegraphics[width=\textwidth]{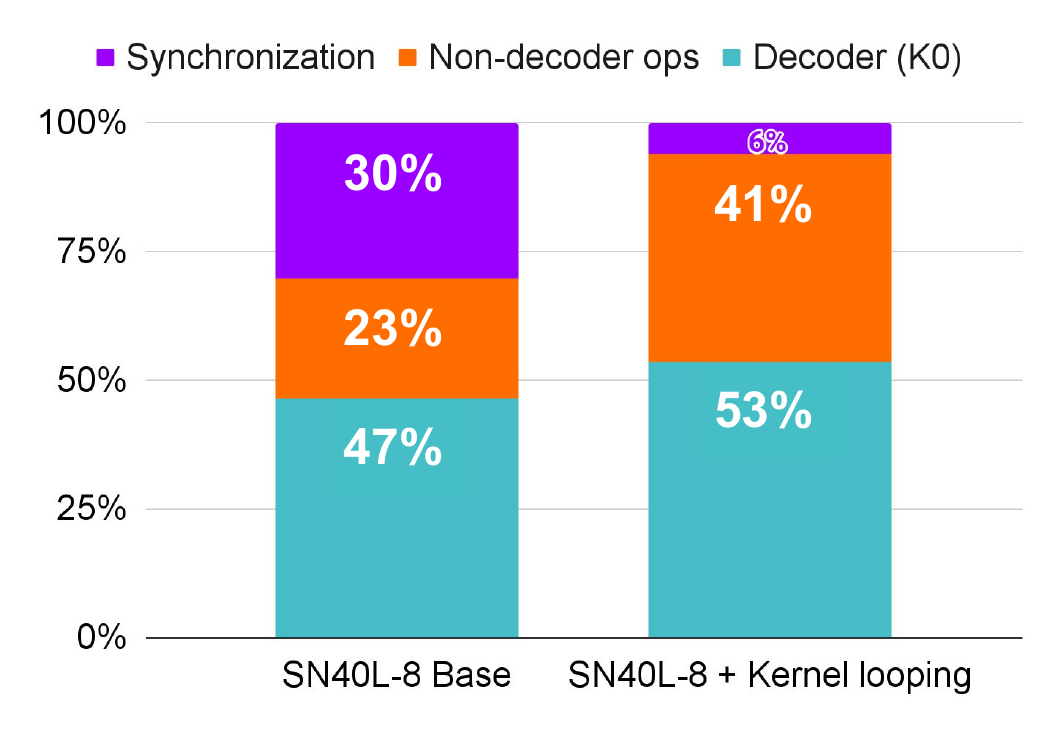}
         \vspace{-15pt}
         \caption{\ts-8 TPOT breakdown by percentage.}
         \label{fig:sn40l-breakdown-percent}
     \end{subfigure}   
        \caption{Time-per-Output-Token (TPOT) breakdown for Llama3.1-8B on \ts-8. Synchronization overheads washes out the performance upside with a single kernel decoder. This paper significantly reduces synchronization overheads to give close to peak TPOT.
        }
        \label{fig:sn40l-breakdown}
\end{figure}

We next characterize the mapping and performance of a decoder layer on \ts. Figure~\ref{fig:decoder-schedule}(ii) shows the kernel schedule to produce one output token. Unlike H100, \ts \space invokes a single kernel $K0$ to execute a decoder. This speaks to a well-studied advantage of dataflow architectures over more traditional data-parallel mappings: the number of operations that can be fused into a single kernel is very flexible. On one extreme, a single logical operation can be spread across the entire chip using data parallelism. On the other extreme, hundreds of operations can be fused into a single kernel via aggressive tiling and pipelining transformations. As we later show in Section~\ref{sec:evaluation}, this characteristic of dataflow architectures, and \ts \space in particular, makes aggressive kernel fusion applicable across a wide range of model sizes and complexities.


By reducing the number of kernel calls, the single decoder mapping reduces the number of synchronization required. However, Figure~\ref{fig:sn40l-breakdown} shows that \textbf{\emph{a single decoder kernel by itself is insufficient to extract peak performance on \ts}}. Synchronization overheads at the end of each call to $K0$ account for more than 30\% of the TPOT, nullifying the performance upside of layer fusion. 

However, a single decoder kernel exposes an optimization opportunity to the compiler: the same kernel is called repeatedly in a chained fashion, where the output of a previous call feeds the input of the next call. The repeated calls stems from the repetitive layer structure inherent in Transformer~\cite{transformer} model architectures, which form the backbone of the most powerful text and multi-modal models in the community today. We therefore propose a generic compiler transformation to rewrite a sequence of such repetitive kernel calls in a schedule into a \emph{single kernel call}. 

Figure~\ref{fig:decoder-kernels}(iii) shows the impact of this transformation on the kernel call schedule. All 32 calls to $K0$ are combined into a single call:  \texttt{all_decoders_nosync(x_in)}. Figure~\ref{fig:sn40l-breakdown} shows that this transformation reduces the synchronization cost by almost 8$\times$. However, there is also a bigger, more profound impact with this transformation: kernel warmup cost is only paid \emph{once} at the beginning of the first decoder. HBM bandwidth stays continuously utilized to stream weights and KV caches in dataflow fashion, fully overlapping it with compute and \emph{allreduce}. Consequently, Llama3.1-8B token generation can achieve 78\% of the peak roofline performance with \ts-8, and scales well to \ts-16 with 76\% of roofline performance as shown in Figure~\ref{fig:gpu-sucks-intro}.

\section{System Overview}
\label{sec:system}
In this section, we describe the implementation details of the kernel looping optimization and provide several simplified examples to help explain its behavior.

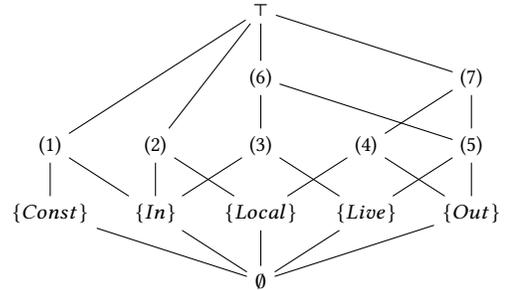
\begin{figure}
\centering
\begin{subfigure}{\columnwidth}
\centering\footnotesize
\begin{tabular}{l|l}\hline 
\textbf{$Const$} & $|A_i| = 1$ \\
                 & Exactly one unique argument across all calls. \\ \hline 
\textbf{$In$}    & $|Reads(p_i)| > 0$ \\ 
                 & Parameter is read at least once. \\ \hline
\textbf{$Local$} & $In(p_i) \cap \exists! j : i \neq j, \forall k \in [0, N-1), a_{j,k} = a_{i,k+1}$ \\
                 & Read parameter: written by $p_j$ in previous call. \\ 
                 &  \\
                 & $Out(p_i) \cap \exists! j : i \neq j, \forall k \in [0, N-1), a_{i,k} = a_{j,k+1}$ \\
                 & \hspace{30pt} $\cap$ $\forall k \in [0, N-1), h\neq j, a_{i,k} \neq a_{h,k+1}$ \\
                 & \hspace{30pt} $\cap$ $\cup WriteSpace(p_j) \geq \cup ReadSpace(p_j)$ \\
                 & Written parameter: read only by $p_j$ in next call \\ 
                 & and write space entirely covers read space. \\ 
                 \hline 
\textbf{$Live$}  & $Out(p_i) \cap \exists j \in [0,N-1) : a_{i,j} \in LiveAfter(c_{N-1})$ \\
                 & At least one stored argument is live after last call. \\ \hline 
\textbf{$Out$}   & $|Writes(p_i)| > 0$ \\
                 & Parameter is written at least once. \\ \hline 
\end{tabular}
\caption{Individual kernel parameter usage properties for the $i'th$ kernel parameter $p_i$ as determined by the corresponding set of arguments $A_i = \{a_{i,0}...a_{i,N-1}\}$ used by kernel calls $c_0$ through $c_{N-1}$. }
\label{fig:argkinds-table}
\end{subfigure}

\begin{subfigure}{\columnwidth}
\centering\footnotesize
\begin{tikzpicture}[x=1.4cm,y=1.8cm]%
\node at (0,0)       (I)   {$\emptyset$};
\node at (-2,0.5)    (C1)  {$\{Const\}$};
\node at (-1,0.5)    (C2)  {$\{In\}$};
\node at (0,0.5)     (C3)  {$\{Local\}$};
\node at (1,0.5)     (C4)  {$\{Live\}$};
\node at (2,0.5)     (C5)  {$\{Out\}$};
\node at (-2,1)      (B1)  {$(1)$}; 
\node at (-1,1)      (B2)  {$(2)$}; 
\node at (0,1)       (B3)  {$(3)$}; 
\node at (1,1)       (B4)  {$(4)$}; 
\node at (2,1)       (B5)  {$(5)$}; 
\node at (0, 1.5)    (A1)  {$(6)$}; 
\node at (2, 1.5)    (A2)  {$(7)$}; 
\node at (0,2)       (T)   {$\top$};

\foreach\i in {1,2,3,4,5}
    \draw (I) -- (C\i);
\draw (C1) -- (B1);
\foreach\i in {1,2,3}
    \draw (C2) -- (B\i);
\foreach\i in {4,5}
    \draw (C5) -- (B\i);
\foreach\i in {2,4} 
    \draw (C3) -- (B\i);
\foreach\i in {3,5} 
    \draw (C4) -- (B\i);

\draw (B3) -- (A1);
\draw (B5) -- (A1);

\draw (B4) -- (A2);
\draw (B5) -- (A2);

\draw (B1) -- (T);
\draw (B2) -- (T);
\draw (A1) -- (T);
\draw (A2) -- (T);

\end{tikzpicture}
\caption{Lattice over sets of usage properties. The lattice follows a constrained form of set intersection/union where, above a single element, only elements (1) -- (7) are well-defined.  }
\label{fig:argkinds-lattice}
\end{subfigure}

\caption{Definitions of kernel parameter usage properties used in kernel looping pattern matching.}
\label{fig:argkinds}
\end{figure}

\subsection{Pattern Matching}
The first step in kernel looping is to pattern match against a \emph{compatible} contiguous list of kernel calls. This is done as a dataflow analysis which determines whether all kernel parameters fit strict patterns of usage across these calls. 
Given a kernel with $M$ parameters $[{p_0, \ldots, p_{M-1}}]$, a candidate set of $N$ calls to this kernel $C=[c_0, \ldots, c_{N-1}]$, where arguments to the $i$'th parameter $p_i$ are $A_i = [a_{i,0},\dots,a_{i,N-1}]$, relevant
usage properties are defined in Figure~\ref{fig:argkinds-table}. 
We assume basic usage analysis ($Reads$, $Writes$) has been run within the kernel body and that a basic liveness analysis ($LiveAfter$) has already been run across kernel calls. 

The $Local$ property is used to match on chained calls like the chained decoder case shown in Figure~\ref{fig:decoder-schedule}. As we will discuss in Section~\ref{sec:system-transform}, this pattern can be used to promote intermediate off-chip tensors to an on-chip buffer. However, doing so requires that several restrictive conditions are met. First, the parameters must be uniquely chained. That is, within $C$, the chained output tensor must only be used by the next call, and the output parameter must be consistently used by the same input parameter.
Additionally, the write address space ($WriteSpace$) over all writes must entirely cover the read address space ($ReadSpace$) to ensure that the read is only dependent on the preceding call. \footnote{Depending on the level of abstraction used by the intermediate representation, proving this condition in general may require polyhedral analysis of access patterns. A compiler using a sufficiently high level domain specific intermediate representation of tensor operations, however, can match on common cases with relative ease.} 


\begin{dmath}
P(p) = Const(p) \vee In(p) \vee Local(p) \vee Live(p) \vee Out(p)
\label{eq:argkinds-equation}
\end{dmath}

Equation~\ref{eq:argkinds-equation} defines the set of usage properties $P$ for a parameter $p$. 
In this equation, the function $F(p)$ yields the singleton usage property set $\{F\}$ if the corresponding condition in Figure~\ref{fig:argkinds-table} holds and $\emptyset$ otherwise. The meet function $\vee$ is defined by the lattice in Figure~\ref{fig:argkinds-lattice}. The candidate call list $C$ then pattern matches successfully if Equation~\ref{eq:argkinds-cond} holds: 
\begin{equation}
Match(C) = \forall i \in [0,M) : P(p_i) < \top 
\label{eq:argkinds-cond}
\end{equation}
We require that $P(p)$ is strictly less than the top element ($\top$), as the top element is considered to be a match failure. All other elements in the lattice are successful matches.

The matching function in Equation~\ref{eq:argkinds-cond} can be used to build an analysis over all kernel calls in the program. In our experiments described in Section~\ref{sec:evaluation}, we found that a basic greedy algorithm was sufficient to fuse and pipeline chains of decoder calls in the targeted language models.
In applications with more complex kernel call interactions and orderings, it may be necessary to do a cost-driven search to maximize the number of optimized contiguous calls.

\subsection{Transformation}
\label{sec:system-transform}

\lstdefinelanguage{transforms}{
  keywords={for,when,otherwise,map,stmts,@onchip},
  keywordstyle=\bfseries,
  identifierstyle=\itshape,
  commentstyle=\color{gray},
  stringstyle=\color{green},
  morecomment=[l]{//},
  morestring=[b]",
  basicstyle=\footnotesize\ttfamily,
  literate={->}{$\rightarrow$}1
}

\begin{figure}
\footnotesize
\vspace{10pt}
\begin{lstlisting}[language=transforms,mathescape]
(1) t(kernel) -> kernel'(t(p$_0$), ..., t(p$_{M-1}$))
                  buffers([p$_0$, ... p$_{M-1}$])
                  for(n = 0; n < N; ++n)
                    t(body(kernel))
                    istores([p$_0$, ... p$_{M-1}$])
                  estores([p$_0$, ... p$_{M-1}$])

(2) t(p : [S]) -> (p : [N x S]) when Group(p)
               -> (p : [S]) otherwise

(3) buffers(ps) -> map buffers ps
(4) buffers(p$_i$) ->  when $P(p_i) == \{In,Local\}$ {
                    $j : \forall k \in [0, N-1), a_{j,k} = a_{i,k+1}$
                    @onchip buf : type(p$_i$) = p$_i$ 
                    buffer[p$_i$] = buf
                    buffer[p$_j$] = buf
                  }

(5) t(body) -> map t stmts(body)

(6) t(p) -> (buffer[p]) when $Local \in P(p)$ 
         -> p[n] when Group(p)
         -> p otherwise

(7) istores(ps) -> map istores ps
(8) istores(p$_i$) ->  when $P(p) == \{Local,Live,Out\}$ {
                    p$_i$[n] = (buffer[p$_i$])
                  }
(9) estores(ps) -> map estores ps
(10) estores(p$_i$) ->  when $P(p) == \{Local,Out\}$ {
                     p$_i$ = (buffer[p$_i$])
                   } 

          
\end{lstlisting}
\vspace{5pt}
\caption{Transformation rules for kernel looping when optimizing a kernel over a list of $N$ calls. The expression $p : [S]$ represents a typed kernel parameter where the tensor shape $S$ is encoded in the type system.}
\label{fig:transform-rules}
\end{figure}

Figure~\ref{fig:transform-rules} summarizes the transformation rules when applying kernel looping on a kernel $k$ over a list of $N$ pattern matched calls. As shown in rule~\ref{fig:transform-rules}(1), this transformation creates a new kernel with a loop, where the number of iterations is equal to the number of calls being optimized. 

In order to use this loop to iterate over arguments across calls, we can view the number of calls $N$ as a new, outermost dimension created by concatenating and reshaping the tensor arguments. We refer to this transformation as ``grouping'', where arguments to a parameter should become a group if Equation~\ref{eq:grouped-cond} holds. Correspondingly, rule~\ref{fig:transform-rules}(2) alters parameter shapes, assumed here to be encoded in the type system.
\begin{equation}
Group(p) = Live\in P(p) \cup (Local\notin P(p)\cap Const\notin P(p))
\label{eq:grouped-cond}
\end{equation}

Due to their chained nature, $Local$ kernel parameters are always found in pairs: an input (read) parameter $p_i$ and an output (written) parameter $p_j$. In rule~\ref{fig:transform-rules}(3 -- 4), each pair of $Local$ parameters introduces a memory allocation initialized by the read parameter, $p_i$. This memory is used as a buffer to store intermediate results across outer loop iterations. Note that double buffering may be required to prevent write after read hazards. For brevity, details of detecting this case are elided from Figure~\ref{fig:transform-rules}, but an example of the transformation with double buffering can be found in Figure~\ref{fig:opt-example-d}.

In rule~\ref{fig:transform-rules}(4), we unconditionally use the $@onchip$ annotation to note that the memory is to be allocated in on-chip accelerator memory, not main memory. In practice, a compiler would likely wish to consult performance and memory capacity models to determine whether this array can or should be in on-chip memory.

The body of the loop introduced in rule ~\ref{fig:transform-rules}(1) is then generated by transforming each statement in the original kernel body via rule~\ref{fig:transform-rules}(5). Rule~\ref{fig:transform-rules}(6) alters any reference to a grouped parameter to index into its outer dimension at outer loop iterator $n$. For example, a reference to parameter $p$ becomes $p[n]$, $p[i]$ becomes $p[n][i]$, and so on. Likewise, a reference to a $Local$ parameter becomes a reference to the associated on-chip memory created by rule~\ref{fig:transform-rules}(4). Uses of $Const$ parameters are unchanged, though later compiler passes may choose to cache data rather than rereading it from off-chip memory. 

Lastly, using rules~\ref{fig:transform-rules}(7 -- 10), on-chip memories are stored back to their associated output parameter. If the output parameter is $Live$, the store occurs at the end of each iteration of the introduced loop. Otherwise, it occurs once at the end of the transformed kernel.


Outside of the kernel, the calls in $C$ are collapsed to a single call. The arguments to this merged call are such that $Const$ parameters use their single, unique argument, $\{Local, In\}$ parameters receive the first argument, and $\{Local,Out\}$ parameters receive only the last argument. Each grouped parameter $p_i$, including $\{Local,Live,Out\}$, receives a grouping of \emph{all} arguments to that parameter, expressed as $\{a_{i,0} ... a_{i,N-1}\}$. 

Grouping can be implemented as a ``ragged'' array of pointers, where corresponding loads and stores jump to different tensor offset addresses with each value of loop iterator $n$. As an optimization, if the compiler has direct control over tensor addresses in main memory, grouped tensors can be organized such that they are linear in memory to maximize spatial locality. However, such groupings cannot be done if the same tensor would appear in multiple disjoint groups.

\lstdefinelanguage{kernelcode}{
  keywords={if, then, else, end, for, while, do, kernel, program, @onchip},
  keywordstyle=\bfseries,
  identifierstyle=\itshape,
  commentstyle=\color{gray},
  stringstyle=\color{green},
  morecomment=[l]{//},
  morestring=[b]"
  basicstyle=\footnotesize\ttfamily
}

\begin{figure*}[h]
\begin{subfigure}{\textwidth}
\centering
\begin{tabular}{p{0.34\textwidth}|p{0.4\textwidth}|p{0.3\textwidth}}
\textbf{Original Program} & \textbf{Transformed Program} & \textbf{Usage Properties} \\ 
 & & \\
\begin{lstlisting}[language=kernelcode]
kernel_a(x[N], y[N], z[N])
  z += x * y

program example_a()
  a[N], b[N], c[N], d[N], e[N]
  call0 = kernel_a(a, b, c)
  call1 = kernel_a(a, e, f)
\end{lstlisting} & 
\begin{lstlisting}[language=kernelcode]
kernel_a'(x'[N], y'[2xN], z'[2xN])
  for(n = 0; n < 2; ++n)
    z'[n] += x' * y'[n]

program example_a'()
  a[N], b[N], c[N], d[N], e[N]
  kernel_a'(a, {b,e}, {c,f})
\end{lstlisting} &
\begin{lstlisting}[language=kernelcode]
LiveAfter(call1)
= {c, f}

P(x) = {Const,In}
P(y) = {In}
P(z) = {In,Live,Out}
\end{lstlisting} \\
\end{tabular}
\caption{Example of optimizing two consecutive but otherwise independent calls to an element-wise multiply-accumulate kernel over tensors of shape \texttt{N}. The transformed kernel iterates over inputs and outputs. No arrays are promoted to on-chip memory.}
\label{fig:opt-example-a}
\end{subfigure}

\vspace{10pt}

\begin{subfigure}{\textwidth}
\centering
\begin{tabular}{p{0.34\textwidth}|p{0.4\textwidth}|p{0.3\textwidth}}
\begin{lstlisting}[language=kernelcode]
kernel_b(x[N], y[N], z[N])
  z = y - x

program example_b()
  a[N], b[N], c[N], d[N]
  e[N], f[N], g[N]
  call0 = kernel_b(a, b, c)
  call1 = kernel_b(c, d, e)
  call2 = kernel_b(f, e, g)
\end{lstlisting} & 
\begin{lstlisting}[language=kernelcode]
kernel_b'(x'[3xN], y'[3xN], z'[3xN])
  for(n = 0; n < 3; ++n)
    z'[n] = y'[n] - x'[n]

program example_b'()
  a[N], b[N], c[N], d[N]
  e[N], f[N], g[N]
  kernel_b'({a,c,f}, {b,d,e}, {c,e,g})
\end{lstlisting} &
\begin{lstlisting}[language=kernelcode]
LiveAfter(call2)
= {c, e, g}

P(x) = {In}
P(y) = {In}
P(z) = {Live,Out}
\end{lstlisting} \\
\end{tabular}
\caption{A program which has ``chained'' calls, but chaining is not consistent across parameters. As with example~\ref{fig:opt-example-a}, arguments are only grouped.}
\label{fig:opt-example-b}
\end{subfigure}

\vspace{10pt}
\begin{subfigure}{\textwidth}
\centering
\begin{tabular}{p{0.34\textwidth}|p{0.4\textwidth}|p{0.3\textwidth}}
\begin{lstlisting}[language=kernelcode]
kernel_c(w[MxM], x[MxN], z[MxN])
  z = w * x

program example_c()
  w0[MxM], w1[MxM]
  x[MxN], y[MxN], z[MxN]
  call0 = kernel_b(w0, x, y)
  call1 = kernel_b(w1, y, z)
\end{lstlisting} & 
\begin{lstlisting}[language=kernelcode]
kernel_c'(w'[2xMxM], x'[MxN], z'[MxN])
  @onchip buf[MxN] = x'
  for(n = 0; n < 2; ++n)
     buf = w[n] * buf
  z' = buf

program example_c'()
  w0[MxM], w1[MxM]
  x[MxN], y[MxN], z[MxN]
  kernel_c'({w0, w1}, x, z)
\end{lstlisting} &
\begin{lstlisting}[language=kernelcode]
LiveAfter(call1)
= {z}

P(w) = {In}
P(x) = {In,Local}
P(z) = {Local,Out}
\end{lstlisting} \\
\end{tabular}
\caption{Example of optimizing two ``chained'' calls to a matrix multiplication kernel of tensors of shape \texttt{MxM} and \texttt{MxN}. The intermediate result $y$ in the chain is not live outside these calls and so is promoted to an exclusively on-chip array \texttt{buf}, denoted by the annotation \texttt{@onchip}.}
\label{fig:opt-example-c}
\end{subfigure}

\vspace{10pt}
\begin{subfigure}{\textwidth}
\centering
\begin{tabular}{p{0.34\textwidth}|p{0.4\textwidth}|p{0.3\textwidth}}
\begin{lstlisting}[language=kernelcode]
kernel_d(w[MxM], x[MxN], z[MxN])
  for (t = 0; t < M; t += T)
    z[t::t+T] = w[t::t+T] * x
   
program example_d()
  w0[MxM], w1[MxM], w2[MxM]
  x[MxN], y[MxN], z[MxN], r[MxN]
  call0 = kernel_d(w0, x, y)
  call1 = kernel_d(w1, y, z)
  call2 = kernel_d(w2, z, r)
\end{lstlisting} & 
\begin{lstlisting}[language=kernelcode]
kernel_d'(w'[3xMxM], x'[MxN], z'[3xMxN])
  @onchip buf0[MxN] = x'
  @onchip buf1[MxN]
  for(n = 0; n < 3; ++n)
    for (t = 0; t < M; t += T)
      buf1[t::t+T] = w[n][t::t+T] * buf0
    buf0 = buf1
    z'[n] = buf1

program example_d'()
  w0[MxM], w1[MxM], w2[MxN]
  x[MxN], y[MxN], z[MxN], r[MxN]
  kernel_d'({w0, w1, w2}, x, {y, z, r})
\end{lstlisting} &
\begin{lstlisting}[language=kernelcode]
LiveAfter(call2)
= {y, z, r}

P(w) = {In}
P(x) = {In,Local}
P(z) = {Local,Live,Out}
\end{lstlisting} \\
\end{tabular}
\caption{An expanded version of example~\ref{fig:opt-example-b}, where there are three ``chained'' calls to a matrix multiplication kernel with tiling factor \texttt{T}.}
\label{fig:opt-example-d}
\end{subfigure}

\caption{Examples of the kernel looping optimization. The left column shows a subset of the original program. The middle column shows the corresponding transformed subset. The right column shows live values after the last shown call and each kernel parameter's usage properties $P$ (see Figure~\ref{fig:argkinds} and Equation~\ref{eq:argkinds-equation}). 
Multi-dimensional shapes are shown in brackets.}
\label{fig:opt-examples}
\end{figure*}

\subsection{Examples}

Figure~\ref{fig:opt-examples} shows several examples of the application of kernel looping. In example~\ref{fig:opt-example-a}, kernel looping is applied to two consecutive but independent (unchained) calls to a kernel performing an element-wise multiply-accumulate. The first kernel parameter \texttt{x} has the same argument \texttt{a} for both calls, so it remains unchanged. Later compiler optimizations may choose to cache this tensor on-chip or re-load it from off-chip memory on each iteration. Arguments to parameters \texttt{y} and \texttt{z} differ across iterations but are entirely independent, so the corresponding arguments for these parameters are grouped and iterated over in the loop introduced in \texttt{kernel_a'}. 

Figure~\ref{fig:opt-example-b} shows an example of a program where calls are not independent: one call's result is input to the following call. However, no parameters meet the criteria for the \texttt{Local} property. Between the first two calls, the result written from parameter \texttt{z} is used as an input to parameter \texttt{x}. But in the last two calls, \texttt{z} is used as an input to \texttt{y}. As a result, arguments are only grouped, like in example~\ref{fig:opt-example-a}. Note that in this case, tensors \texttt{c} and \texttt{e} appear in multiple argument groups. 

In Figure~\ref{fig:opt-example-c}, the program has two ``chained'' calls to a matrix multiplication kernel, where the result of one call is used as an input to the next call. Since the intermediate tensor \texttt{y} is not used except in the second call, this chaining is promoted to exclusively on-chip memory \texttt{buf}. This memory is initialized with the first argument to parameter \texttt{x}, used to store intermediate results, and then used to write the final result to parameter \texttt{z} at the end of the kernel. Parameter \texttt{w} is transformed using the same mechanism as in example~\ref{fig:opt-example-a}, where all arguments are grouped and iterated over. 

Figure~\ref{fig:opt-example-d} expands on example~\ref{fig:opt-example-c}: the matrix multiplication is now tiled on one dimension and all intermediate results are live after the last targeted call. As before, kernel looping introduces an on-chip memory to host the chained intermediate result. However, \texttt{buf0} is reread while the next iteration is being produced, meaning the on-chip buffer needs to be double buffered. Additionally, since all intermediate results are live, the write to \texttt{z} occurs at the end of every outermost iteration rather than at the end of the entire kernel.  
\section{Evaluation}
\label{sec:evaluation}
We evaluate kernel looping across some of the most powerful open source models to date. First, we evaluate its \emph{generality}, or how broadly applicable the optimization is. Second, we evaluate its \emph{scalability} over increasing numbers of \ts \space sockets. Lastly, we compare performance between \ts \space with kernel looping and DGX H100.

\subsection{Methodology}
\begin{table}
\centering\small
\def\arraystretch{1}%
\begin{tabular}{|p{1.4in} |p{0.7in}|p{0.7in}|} \hline
\textbf{Model, Precision} & \textbf{Batch Sizes} & \textbf{Context Lengths} \\ \hline
Llama~\cite{llama-herd} 3.1 8B, BF16    & 1, 8, 16   & 4K, 8K, 16K  \\ \hline
Llama~\cite{llama-herd} 3.1 70B, BF16   & 1, 8      & 4K, 8K       \\ \hline
Llama~\cite{llama-herd} 3.1 405B, BF16  & 1, 4       & 4K           \\ \hline
Llama~\cite{llama-herd} 3.2 3B, BF16    & 1, 8, 16   & 4K           \\ \hline
Mixtral~\cite{mixtral} 8x7B, BF16    & 1, 8, 16      & 4K           \\ \hline
Qwen~\cite{qwen2.5} 2.5 72B, BF16    & 1, 8, 16   & 4K           \\ \hline
\end{tabular}
\caption{Model architectures, parameters, precision, batch sizes, and sequence lengths evaluated in this paper.}
\label{tab:benchmarks}
\end{table}

\begin{table}
\centering\small
\def\arraystretch{1}%
\begin{tabular}{|p{0.9in} |p{0.5in}|p{0.7in}|p{0.7in}|} \hline
\textbf{Platform} & \textbf{Number of sockets} & \textbf{Total BF16 TFLOPS} & \textbf{Total HBM Bandwidth} \\ \hline
DGX H100    & 8     & 8000   & 24 TB/s      \\ \hline
\ts-8    & 8     & 5104   & 12.8 TB/s    \\ \hline
\ts-16  & 16    & 10208  & 25.6 TB/s    \\ \hline
\end{tabular}
\caption{TFLOPs and HBM bandwidths of hardware configurations being evaluated. Single socket numbers have been multiplied by the socket count in this table.}
\label{tab:hardware-configs}
\end{table}

\begin{figure*}
        \centering
         \includegraphics[width=0.85\linewidth]{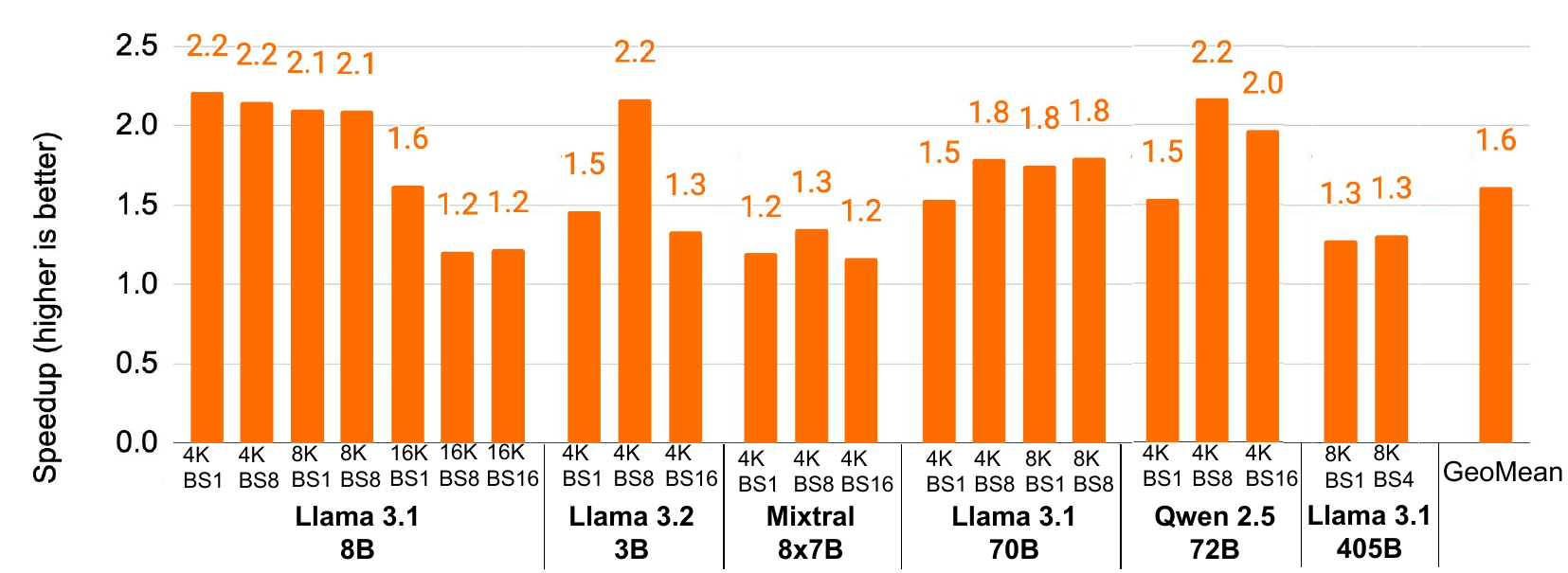}
         \vspace{-10pt}
         \caption{Kernel looping speeds up a broad variety of model architectures, batch sizes, and sequence lengths.}
         \label{fig:eval-generality}
\end{figure*}

\begin{figure*}
        \centering
         \includegraphics[width=0.8\linewidth]{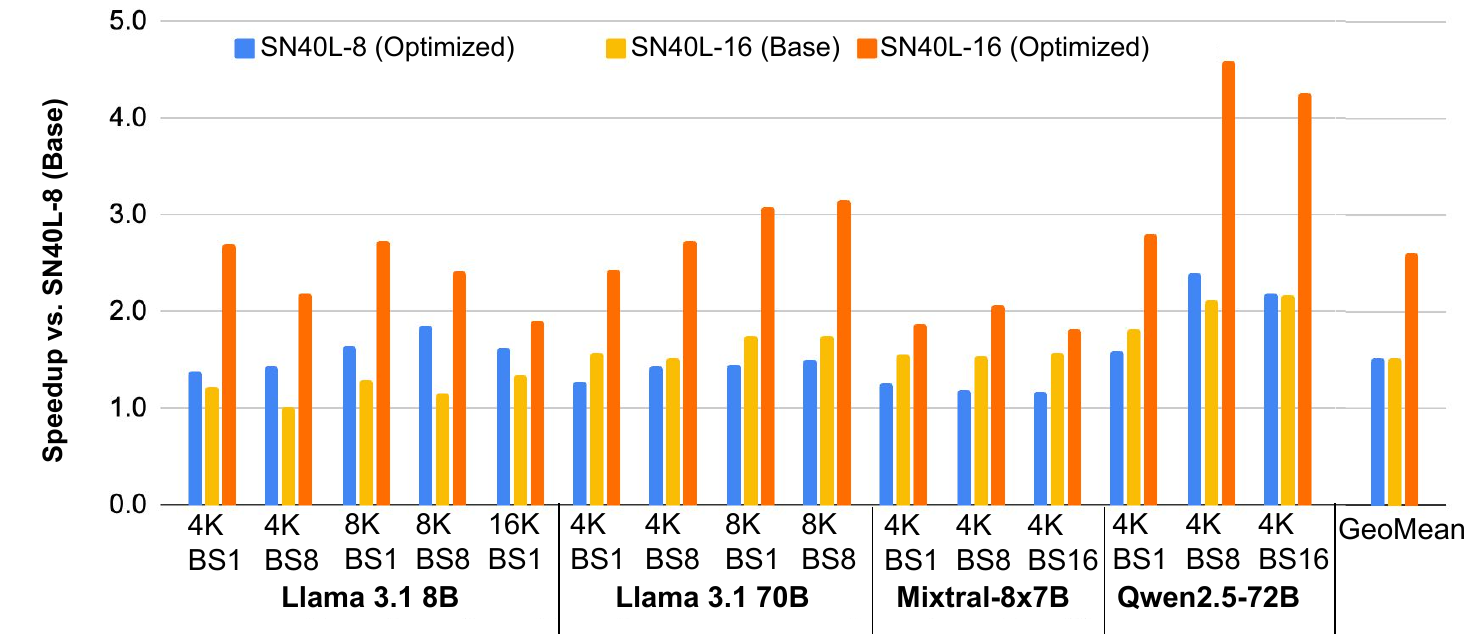}
         \caption{Kernel looping enables performance scaling going from 8 to 16 \ts \space sockets. \ts-16 (Base) without kernel looping can be slower than \ts-8 for smaller models.}
         \label{fig:eval-chip-scaling}
\end{figure*}

Table~\ref{tab:benchmarks} describes the benchmark configurations studied in this paper. Table~\ref{tab:hardware-configs} describes the compute TFLOPS and memory bandwidths of the hardware platforms being evaluated. As this paper focuses only on the decode stage of inference, all evaluations in this section measure and report the decode throughput. All benchmarks are run and measured on real hardware. The throughput number is obtained by measuring the time per output token (TPOT) in the steady state on all platforms. DGX H100 performance numbers are obtained by executing the models using TensorRT-LLM~\cite{tensorrt-llm} and using the included benchmarking scripts that measure steady state token generation throughput. \ts \space performance numbers are obtained by measuring and averaging the tokens/second on 10 prompts that generate a range of output tokens ($<100$ to $>1000$ output tokens). All models targeting \ts \space were compiled with the kernel looping optimization.

\subsection{Generality}
Figure~\ref{fig:eval-generality} shows the speedups obtained on \ts-16 with kernel looping enabled, over a baseline without kernel looping. Kernel looping speeds up a broad variety of model architectures, parameter sizes, batch sizes, and sequence lengths, with a geomean speedup of 1.6$\times$. We highlight that kernel looping is not limited only to smaller models. Figure~\ref{fig:eval-generality} shows that larger models like Qwen2.5-72B observe speedups close to 2$\times$ even at large batch sizes, highlighting the nontrivial impact of synchronization overheads on decode performance without kernel looping.

\subsection{Scalability}

Next, we quantify the impact of kernel looping on scaling performance over multiple sockets. Figure~\ref{fig:eval-chip-scaling} shows the speedups of both three configurations over a baseline \ts-8: \ts-8 with kernel looping, and \ts-16 with and without kernel looping. We observe that the optimization enables performance scaling from 8 to 16 \ts \space sockets, with a geometric mean speedup of 2.5$\times$ over an \ts-8 baseline without kernel looping. We also observe that the performance of models suffer significantly on \ts-16, running slower than \ts-8 due to the out-sized impact of synchronization overheads at the end of each decoder. Kernel looping eliminates these overheads.

\subsection{Performance vs. DGX H100}
\begin{figure}
        \centering
        \includegraphics[width=\linewidth]{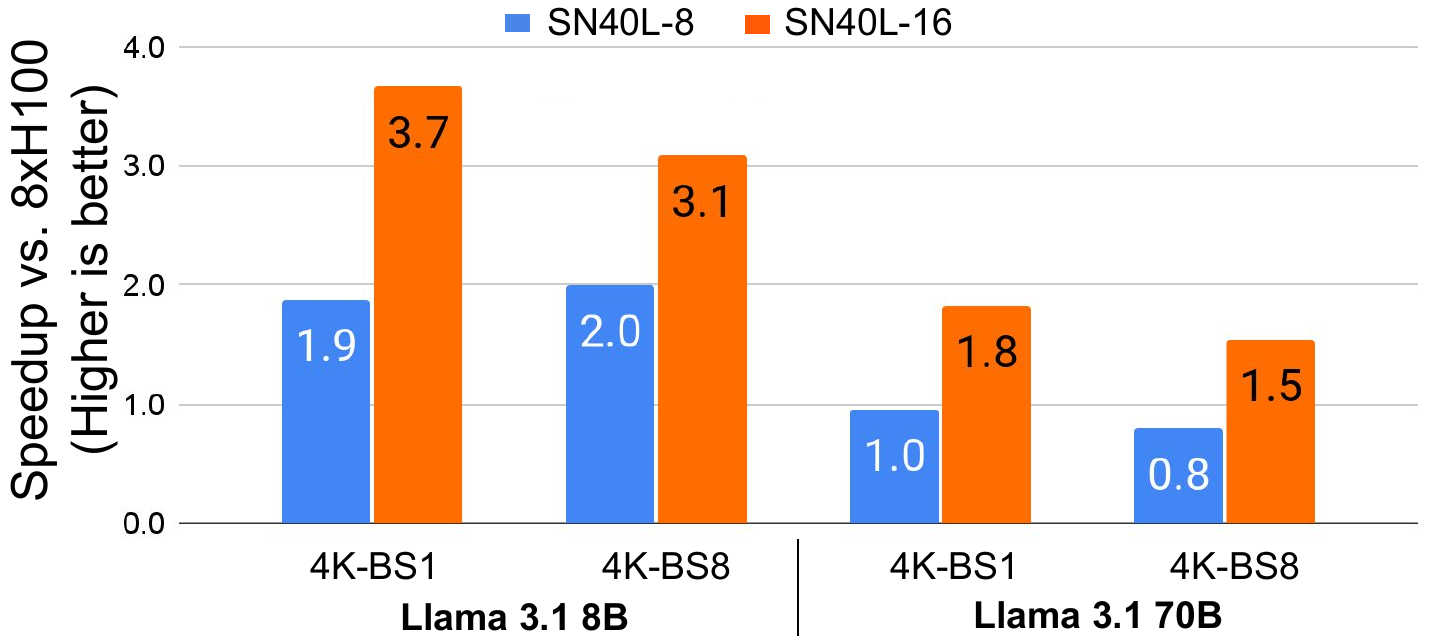}
         \caption{Kernel looping enables higher performance on \ts \space over H100. \ts-8 matches outperforms DGX H100 in spite of having 50\% lower aggregate HBM bandwidth.}
         \label{fig:eval-speedup-gpu}
\end{figure}

\begin{figure}
        \centering
        \includegraphics[width=\linewidth]{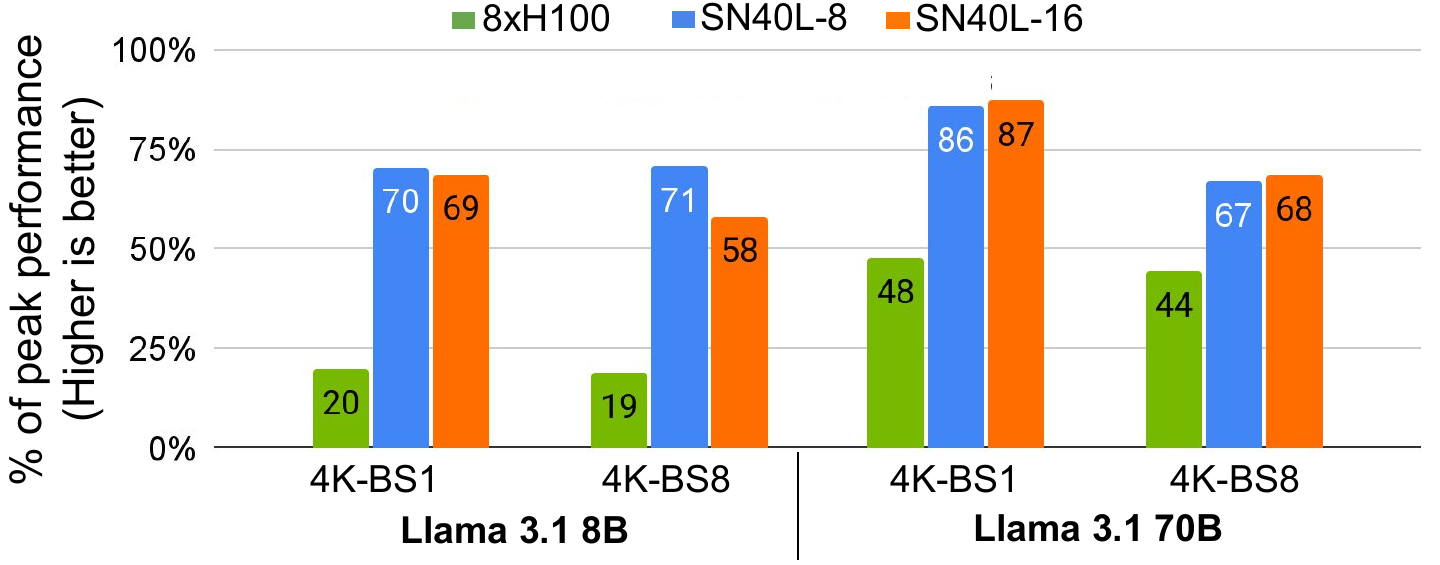}
         \caption{Percentage of peak decode throughput achieved on each platform. With kernel looping, \ts \space achieves 95\% of peak decode throughput on Llama3.1-70B.}
        \vspace{-8pt}
         \label{fig:eval-util-gpu}
\end{figure}

\begin{figure}
        \centering
         \includegraphics[width=\linewidth]{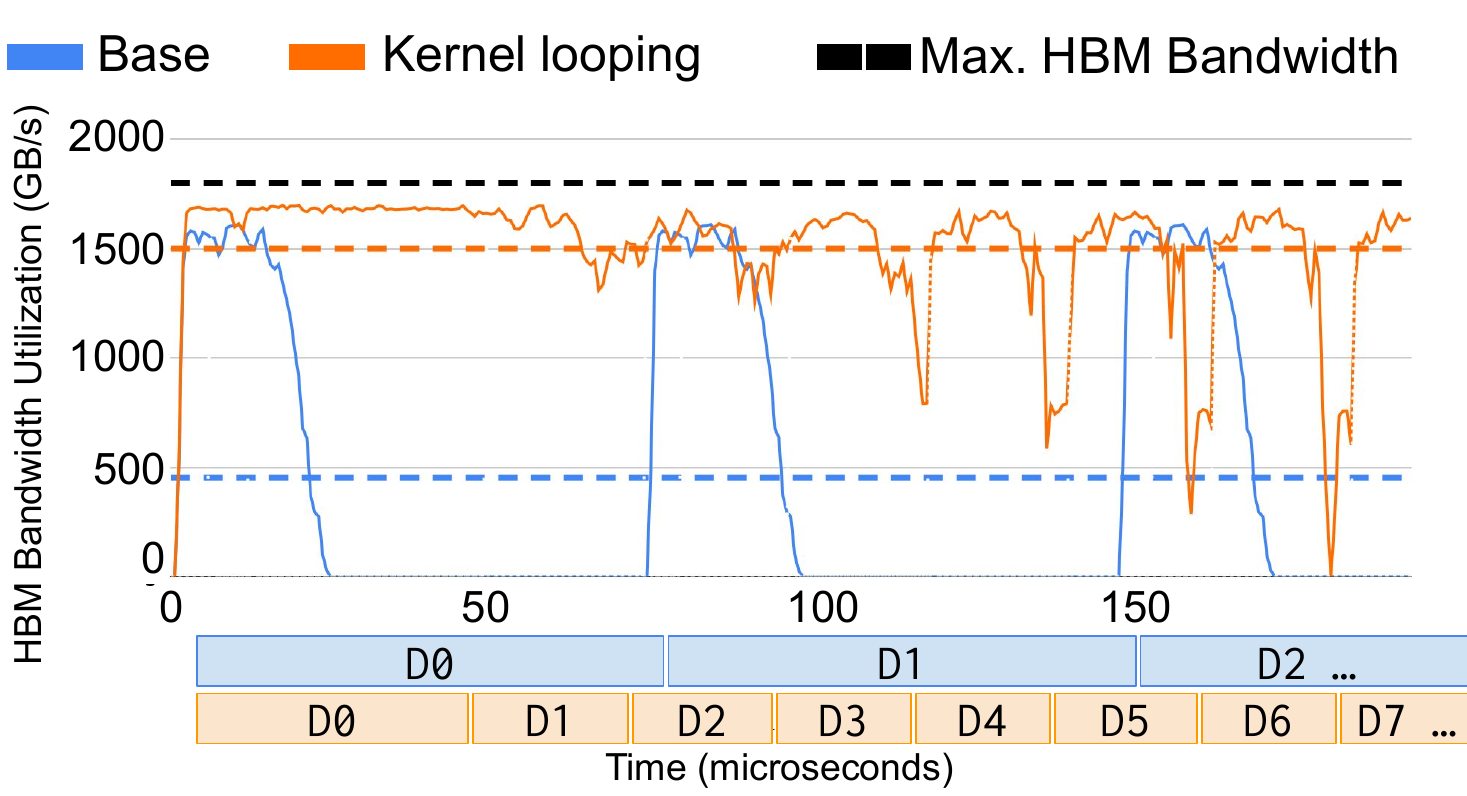}
         \caption{HBM bandwidth trace from one \ts \space socket running Llama3.1-8B. Kernel looping(orange) sustains higher HBM bandwidth utilization over the baseline (blue).}
         \label{fig:eval-hbm-util}
\end{figure}

Figure~\ref{fig:eval-speedup-gpu} shows the speedups achieved by \ts-8 and \ts-16 over 8xH100. \ts-8 matches or outperforms DGX H100 by up to $2\times$ in spite of having only 50\% of the DGX H100's peak HBM bandwidth. A detailed breakdown of performance scaling issues on DGX H100 can be found in Section~\ref{sec:motivation}. While we only compare four model configurations against the DGX H100, we observe that \ts \space with kernel looping achieves similar speedups on the other models, based on publicly benchmarked decode performance numbers from several inference API providers~\cite{aa}.

Figure~\ref{fig:eval-util-gpu} plots the performance as a percentage of peak achievable performance. Here, `peak' (100\%) on each platform is calculated as the time taken to stream the weights and KV cache values at peak HBM bandwidth. The DGX H100 only achieves 21\% of its peak performance as synchronization overheads wash away the upside of higher HBM bandwidth. In contrast, both \ts-8 and \ts-16 achieve over 70\% of peak performance, up to 93\% on Llama3.1-70B.

Figure~\ref{fig:eval-hbm-util} illustrates the benefits of kernel looping with an HBM bandwidth utilization trace obtained from a single socket of a \ts-16 system executing Llama3.1-8B. Kernel looping (orange) sustains a much higher HBM bandwidth utilization compared to the baseline (blue), which shows periodic drops in HBM bandwidth due to synchronization overheads. Consequently, kernel looping enables executing many more decoder layers during the same time window.

Kernel looping and all models studied in this section have been deployed in production.

\section{Related Work}
\label{sec:related}

We discuss one class of related work by casting \loopconversion~as a logical composition of two related transformations - loop rerolling and kernel fusion. 
Loop (re)rolling is done in CPU compilers to reduce code size to target memory-limited environments like embedded devices. The transformation may be performed in an isolated fashion, such as in RoLAG~\cite{loop-rerolling-rolag} and RollBin~\cite{loop-rerolling-rollbin} or to condense intermediate or final code size in a larger software pipelining optimization~\cite{loop-rerolling-urpr}. Loop rerolling is also used to ``rediscover'' unrolled loops when decompiling binary code~\cite{loop-rerolling-hw2} and hardware netlists~\cite{loop-rerolling-hw1}. In the latter case, rerolling serves to both decrease artifact size and speed up hardware simulation time. However, this speedup happens because it exploits optimizations in the targeted hardware simulator related to simulating rolled loops in the hardware description language. In contrast, 
\loopconversion~is agnostic to code size, and aims primarily to improve the performance of the model.  

Kernel fusion is a commonly known optimization which combines multiple smaller kernels into a single larger kernel. Fusion has been shown to improve both GPU energy efficiency~\cite{vertical-kernel-fusion} and performance~\cite{vertical-kernel-fusion2, dnn-fusion}, particularly for memory-bound kernels. 
While kernel fusion typically refers to the \emph{vertical} fusion of chained, data-dependent kernel calls, \emph{horizontal} fusion across independent GPU kernel calls has also been proposed~\cite{horizontal-kernel-fusion} and shown to help improve performance by hiding instruction latencies.


Several methods have been proposed to reduce the runtime overhead of kernel calls. Ismayilov, et. al.,~\cite{gpus-run-orchestration} present an execution model which offloads synchronization and orchestration of GPU kernels in a multi-GPU environment to the GPUs themselves, thus improving performance and reducing communication latency. CUDA Streams~\cite{cuda-streams-concurrency} and CUDA Graphs~\cite{cudagraphs} enable concurrent kernel execution and minimizes host involvement in launching multiple kernels on GPUs. In contrast, kernel looping eliminates the need for multiple calls and overlaps the execution of operations.

\section{Conclusions}
\label{sec:conclusions}
This paper presented kernel looping, a novel compiler optimization to improve inference performance by eliminating synchronization at kernel call boundaries. 
Kernel looping leverages the repetitive structure of layers in LLMs to transform multiple consecutive kernel calls into a single call with a pipelined outer loop, thereby minimizing synchronization points and improving compute-memory overlap. Evaluation on \ts, a commercial reconfigurable dataflow accelerator, shows significant performance improvements with a 3.7$\times$ speedup over DGX H100 with TensorRT-LLM, and scaling efficiently to larger configurations with up to 90\% of peak bandwidth utilization on multi-socket setups. Kernel looping achieves a geometric mean speedup of 1.6$\times$ over a variety of models, demonstrating general applicability. Additionally, kernel looping achieved a 2.5$\times$ geometric mean speedup when scaling from 8 to 16 \ts \space sockets. Kernel looping and the models evaluated in this paper are deployed in a commercial AI inference cloud. Kernel looping establishes a new performance ceiling for inference on dataflow accelerators like \ts. The work lays a foundation for further compiler and hardware co-design to fully exploit the potential of modern AI inference platforms.


\bibliographystyle{plain}
\bibliography{references}

\end{document}